\documentclass[conference]{IEEEtran}

\IEEEoverridecommandlockouts 
\usepackage[english]{babel}

\usepackage{graphicx}
\usepackage{animate}
\usepackage{epsfig} 				
\usepackage{epstopdf}

\usepackage{listings}
\usepackage{color}
\usepackage{nameref}

\usepackage{xcolor}
\usepackage[colorlinks,bookmarks=false,citecolor=blue,linkcolor=blue,urlcolor=blue]{hyperref}
\usepackage[numbers]{natbib}	 			
\usepackage{amsmath}	 			
\usepackage{amssymb}  				

\usepackage{dsfont}			
\usepackage{mathtools}

\usepackage{epigraph}
\usepackage{lscape}
\usepackage[]{nomencl}				
\usepackage{algorithm}
\usepackage{algorithmic}
\usepackage{multicol}
\usepackage{multirow}
\usepackage{makecell}
\usepackage{etoolbox}
\usepackage{graphics}

\usepackage{wrapfig}

\usepackage{siunitx}

\usepackage{floatflt}

\usepackage{url}

\usepackage{placeins}
\usepackage{bm}

\usepackage{booktabs}
\usepackage{textcomp,mathcomp}
\usepackage{bm}
\usepackage{xspace}
\usepackage{mdframed}

\newcommand{\link}[1]{\colora{\url{#1}}}

\newcommand{\ProjectWeb}[0]{\href{https://HiPi-Sensor.github.io
}{https://HiPi-Sensor.github.io}}

\newcommand{\sensor}[0]{HiPi\xspace}

\begin{document}

\title{\LARGE \bf \sensor: Reproducible High-Fidelity Piezoresistive Sensors \\for Robotic Manipulation}
\vspace{-0.5cm}
\author{
Changyi Lin$^{1}$, Raihan Haque$^{2}$, Hui-Ping Wang$^{2}$, Ding Zhao$^{1}$\\
[0.05cm]
$^{1}$Carnegie Mellon University $^{2}$General Motors\\
[0.05cm]
\ProjectWeb
}

\maketitle
\IEEEpeerreviewmaketitle

\begin{abstract}
Piezoresistive tactile sensors are attractive for robotic manipulation because they are thin, lightweight, low-cost, and scalable to dense large-area sensing. However, existing systems still face a practical trade-off: recent reproducible designs emphasize accessibility and ease of reproduction, whereas high-fidelity readout architectures remain more difficult to fabricate, assemble, and deploy. We present \sensor, a reproducible high-fidelity piezoresistive sensing system for robotic manipulation. Building on a low-crosstalk readout principle, \sensor redesigns the complete hardware stack around reproducibility, deployability, and multi-sensor scalability. The system includes a compact readout PCB compatible with commercial PCB fabrication and assembly services, eliminating manual soldering; a smaller and lower-cost STM32-based MCU module; an optimized communication pipeline that achieves $220\,\mathrm{Hz}$ readout in a bimanual setup with four dense tactile arrays ($2048$ taxels in total); and FPCB-based conductive layers that simplify sensor fabrication and stacking. Experiments with structured 3D-printed contact patterns show that \sensor preserves contact geometry substantially better than a reproducible baseline, improving the average IoU from $0.428$ to $0.797$ and the average Dice score from $0.539$ to $0.886$. These results suggest that \sensor bridges an important gap between reproducible fabrication and high-fidelity readout, making dense piezoresistive tactile sensing more practical for bimanual manipulation and multi-fingered robotic systems.
\end{abstract}

\begin{figure}[t]
\centering
\includegraphics[width=\linewidth]{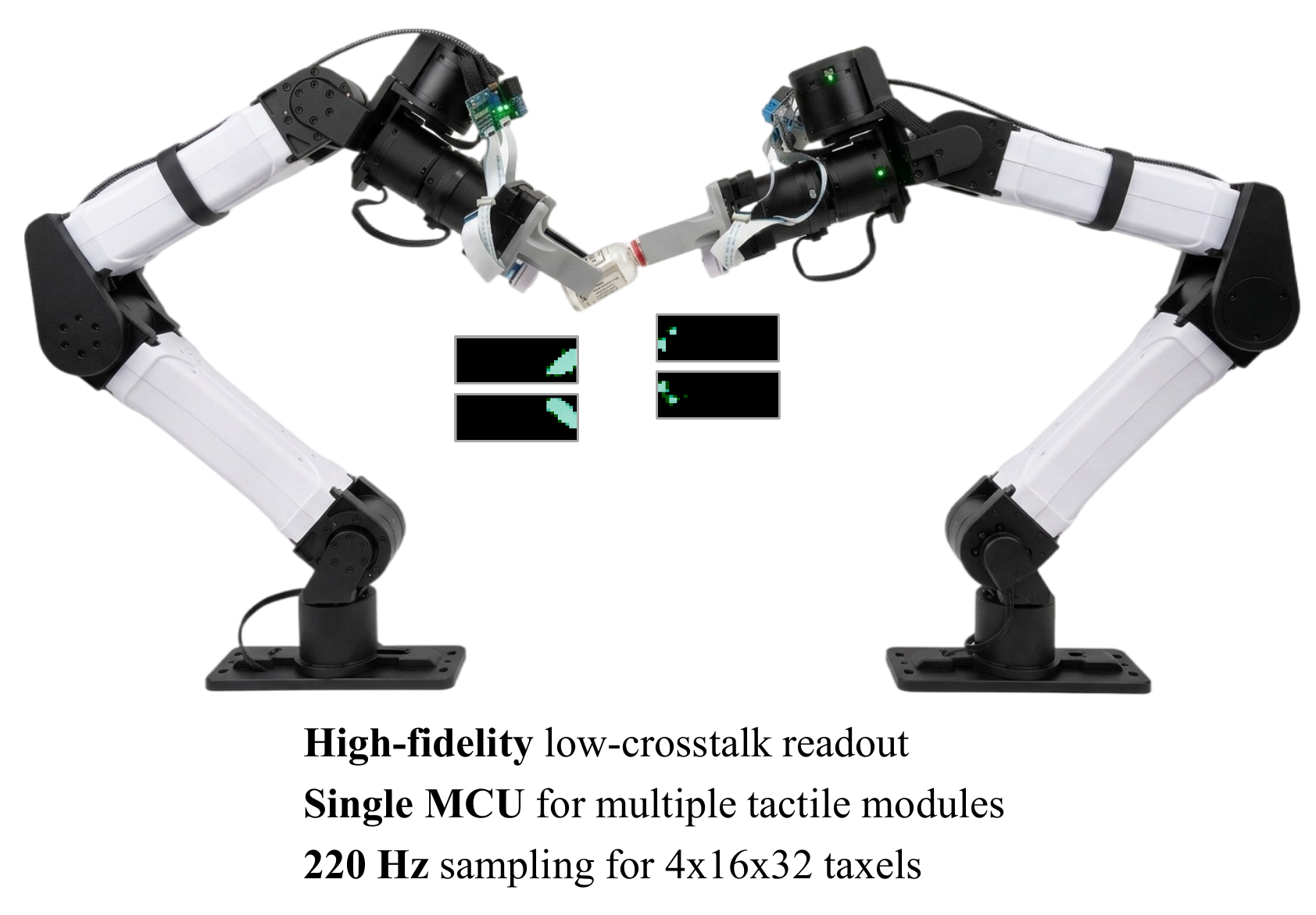}
\vspace{-0.5cm}
\caption{Overview of \sensor in a bimanual robotic manipulation setup. Our system combines high-fidelity low-crosstalk tactile readout, a shared-MCU architecture for multiple tactile modules, and 220\,Hz sampling for four $16\times32$ tactile arrays. Insets show representative tactile images.}  
\label{fig:teaser}
\end{figure}

\section{Introduction}

Tactile sensing provides direct information about physical contact that is difficult to infer reliably from vision alone. As robots increasingly perform contact-rich manipulation and whole-body interaction, tactile sensing is becoming an important component of robotic perception and control. Among tactile sensing modalities, piezoresistive sensors are particularly attractive because they are thin, lightweight, low-cost, and naturally scalable to dense large-area sensing.

These advantages have motivated a growing family of piezoresistive tactile systems. Early work~\cite{sundaram2019learning} established the practicality of dense fabric-based tactile sensing, but reproducing the system still required manual soldering of the readout PCB and manual placement of conductive cables in the sensor. Later works improved different parts of this pipeline. In robotic manipulation,~\cite{huang20243d} simplified the readout electronics to improve accessibility. In quadrupedal interaction,~\cite{lin2025locotouch} improved sensor fabrication through laser-cut conductive fabric. More recently,~\cite{murphy2025fits} introduced an FPCB-based conductive-layer fabrication method, which was later adopted by~\cite{huang2025vt}. As reflected by the recent FlexiTac system~\cite{huang2026flexitac}, this line of work has evolved into one of the most reproducible and accessible piezoresistive tactile solutions for the robot learning community.

In parallel,~\cite{johnson2025scaling} showed that substantially higher-quality tactile signals can be achieved through a low-crosstalk readout architecture for large-scale piezoresistive arrays, establishing a strong baseline for high-fidelity readout. Existing piezoresistive systems have therefore progressed along two complementary directions: reproducible, community-oriented fabrication on one side, and high-fidelity readout on the other. What remains missing is a practical system that combines both.

To address this gap, we present \sensor, a reproducible high-fidelity piezoresistive sensing system for robotic manipulation. \sensor builds on the low-crosstalk readout principle of~\cite{johnson2025scaling}, but redesigns the overall hardware stack with reproducibility and deployability as primary objectives. Specifically, we redesign the tactile readout PCB to be compatible with commercial PCB fabrication and assembly services, thereby eliminating manual soldering. We reduce the PCB footprint from $49\times89\,\mathrm{mm}$ to $32\times62\,\mathrm{mm}$, replace the Arduino Due with a smaller and lower-cost WeAct Black Pill V2.0 based on the STM32F411CEU6, and optimize the communication pipeline to achieve $220\,\mathrm{Hz}$ readout in the demonstrated bimanual setup (Fig.~\ref{fig:teaser}) with four $16\times32$ tactile arrays. For tactile sensor fabrication, we adopt the FPCB-based conductive-layer method of~\cite{murphy2025fits}, which removes the need for manual conductive-cable alignment and simplifies sensor stacking.

\sensor is positioned against two main baselines. Compared with~\cite{johnson2025scaling}, it improves compactness, deployability, and ease of reproduction while preserving the benefits of low-crosstalk readout. Compared with FlexiTac~\cite{huang2026flexitac}, it improves signal fidelity and system scalability through lower crosstalk, faster readout, more efficient communication, cleaner synchronization, and an extensible architecture in which one MCU supports multiple readout PCBs through SPI-based communication. These properties make \sensor well suited for robotic manipulation setups such as bimanual platforms and multi-fingered end-effectors. Overall, this paper does not propose a new sensing principle; instead, it bridges the gap between the reproducibility of recent community-oriented piezoresistive sensors and the signal quality of low-crosstalk readout hardware. To support practical adoption by the community, we will open-source the system upon publication.

\section{System Design}

\begin{figure}[t]
    \centering
    \includegraphics[width=\columnwidth]{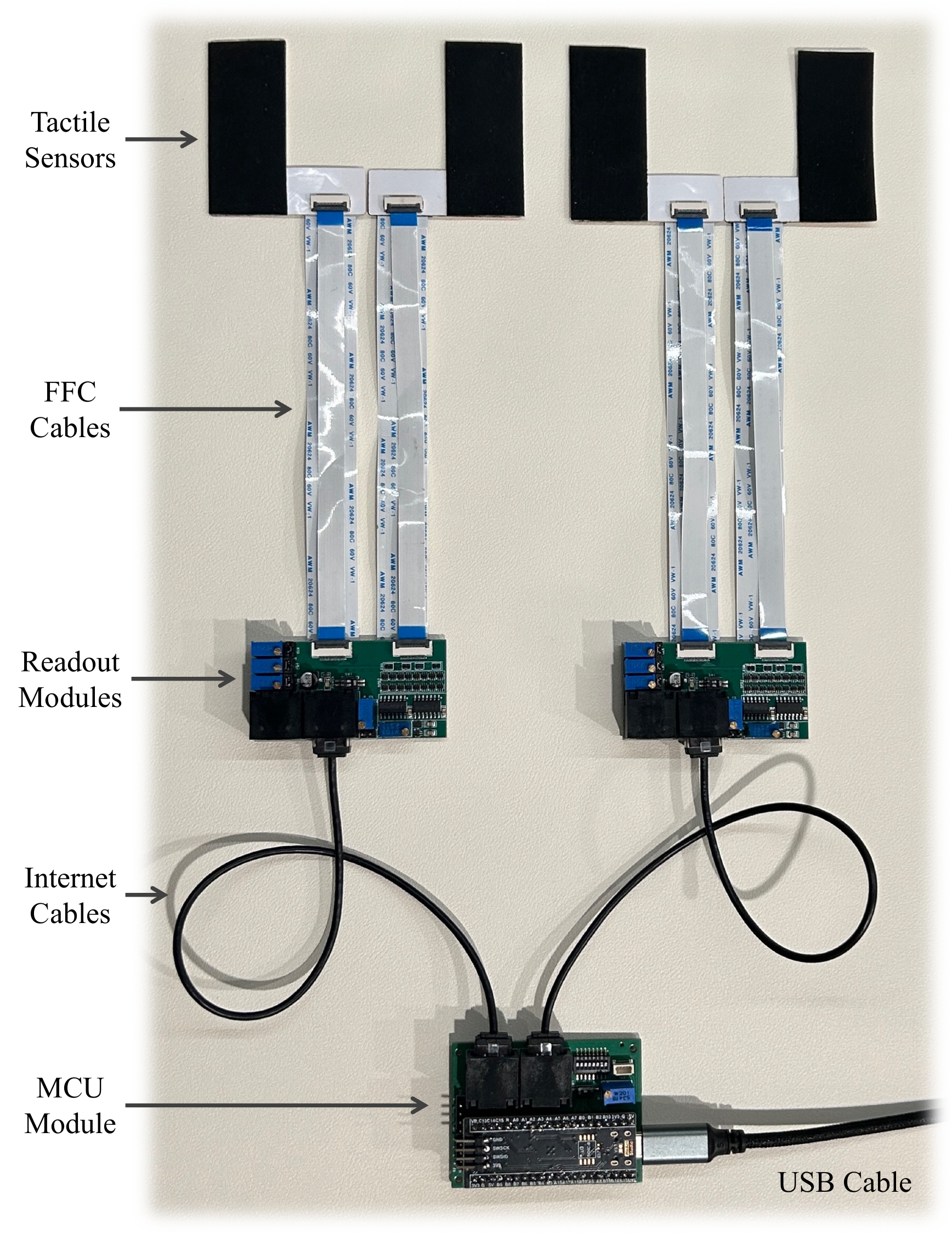}
    \vspace{-0.5cm}
    \caption{Hardware overview of \sensor. Two tactile sensors are connected to one readout module through FFC cables, and two readout modules are connected to a shared MCU module through Ethernet cables. The MCU module communicates with the host computer through USB.}
    \label{fig:overview}
\end{figure}

As shown in Fig.~\ref{fig:overview}, the hardware stack consists of tactile sensors, readout modules, and a shared MCU module. In the demonstrated bimanual setup (Fig.~\ref{fig:teaser}), each gripper uses two tactile sensors, one on each jaw, together with one readout module, while a shared MCU module provides communication and control for both grippers.

\subsection{Hardware Overview}

Each tactile sensor contains a $16\times32$ tactile array over an active area of $25\times60\,\mathrm{mm}$. Two such sensors are mounted on the two jaws of one gripper, resulting in $16\times64$ taxels per gripper. For signal acquisition, the two sensors on each gripper are connected to one dedicated readout module through FFC cables. Therefore, the full bimanual setup contains two readout modules in total, one for each gripper.

At the system level, the two readout modules are connected to a shared MCU module based on the WeAct Black Pill V2.0. To simplify wiring and improve extensibility, the MCU is mounted on a custom carrier PCB that provides two Ethernet connectors, allowing the MCU module to communicate with the two readout modules separately. This design supports a clean bimanual configuration while providing a convenient path to other multi-sensor robotic platforms.

\subsection{Electronics Design}

\begin{figure}[t]
    \centering
    \includegraphics[width=\columnwidth]{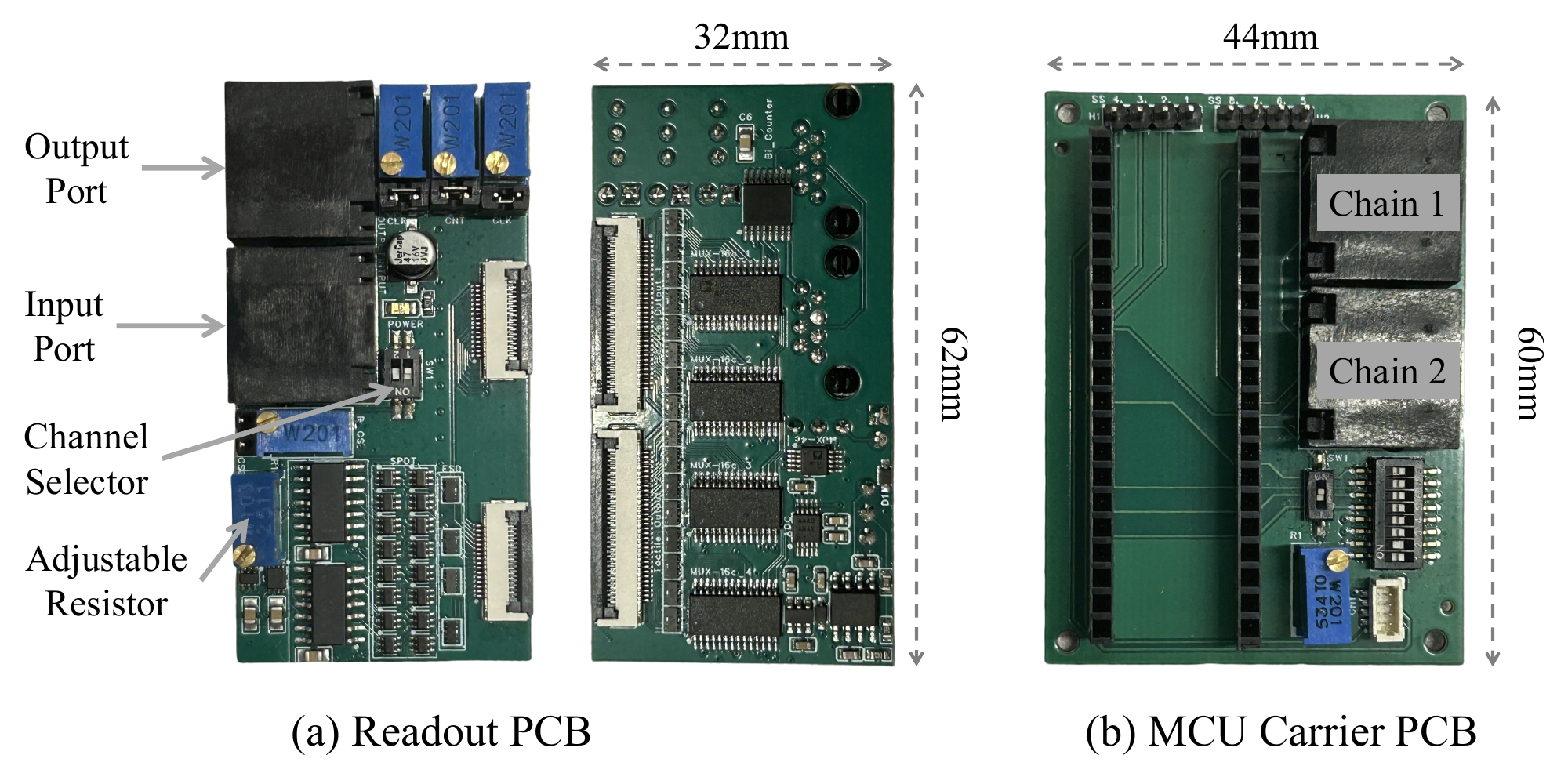}
    \vspace{-0.5cm}
    \caption{Detailed view of the electronics in \sensor. (a) Compact readout PCB with labeled input/output ports, channel selector, adjustable resistor, and board dimensions. (b) MCU carrier PCB used with the WeAct Black Pill V2.0 and Ethernet connectors for communication with readout modules.}
    \label{fig:pcb}
\end{figure}

The electronics stack consists of a compact readout PCB for tactile signal acquisition and an MCU carrier PCB for communication and control, as shown in Fig.~\ref{fig:pcb}. Building directly on the low-crosstalk readout principle of~\cite{johnson2025scaling}, we redesign the readout PCB for practical reproducibility and compact deployment. The PCB is compatible with commercial PCB fabrication and assembly services, eliminating manual soldering. This is important for the robotics community because many users need tactile sensing as a practical research tool rather than as a separate electronics project. In addition, the PCB footprint is reduced by $54.4\%$, from $49\times89\,\mathrm{mm}$ in~\cite{johnson2025scaling} to $32\times62\,\mathrm{mm}$ in our system, making it easier to integrate into robotic platforms with limited available space. The readout PCB also includes an adjustable resistor for voltage-gain control, allowing the same board to adapt to tactile sensors with different force--resistance characteristics.

For the controller, we replace the Arduino Due used in~\cite{johnson2025scaling} with a WeAct Black Pill V2.0 based on the STM32F411CEU6. This change reduces both size and cost while remaining sufficient for embedded tactile acquisition and communication. To support modular multi-sensor deployment, the MCU is mounted on a custom carrier PCB and connected to the readout PCBs through Ethernet cables. Each Ethernet cable carries eight signals: power, ground, two reset/control lines, two SPI lines, and two chip-select lines. On each readout PCB, a channel selector determines which chip-select line is used, while the output port supports extension to additional readout boards. The MCU carrier PCB exposes eight chip-select lines in total, enabling up to four readout PCBs to be connected across two Ethernet chains, with two boards on each chain. Additional readout PCBs can be supported by routing extra chip-select signals through separate single-line cables.

We further optimize the communication pipeline to increase the effective sampling frequency. Since each ADC reading is $12$ bits, we pack every three ADC readings into two $16$-bit words before transmission to the host computer. This reduces communication overhead and contributes to the achieved $450\,\mathrm{Hz}$ readout for two $16\times64$ tactile arrays and $220\,\mathrm{Hz}$ readout for four arrays. Overall, this communication architecture provides a cleaner, more extensible, and better synchronized alternative to tightly coupling one MCU to one tactile readout board.

\subsection{Tactile Sensor Design}

\begin{figure}[t]
    \centering
    \includegraphics[width=\columnwidth]{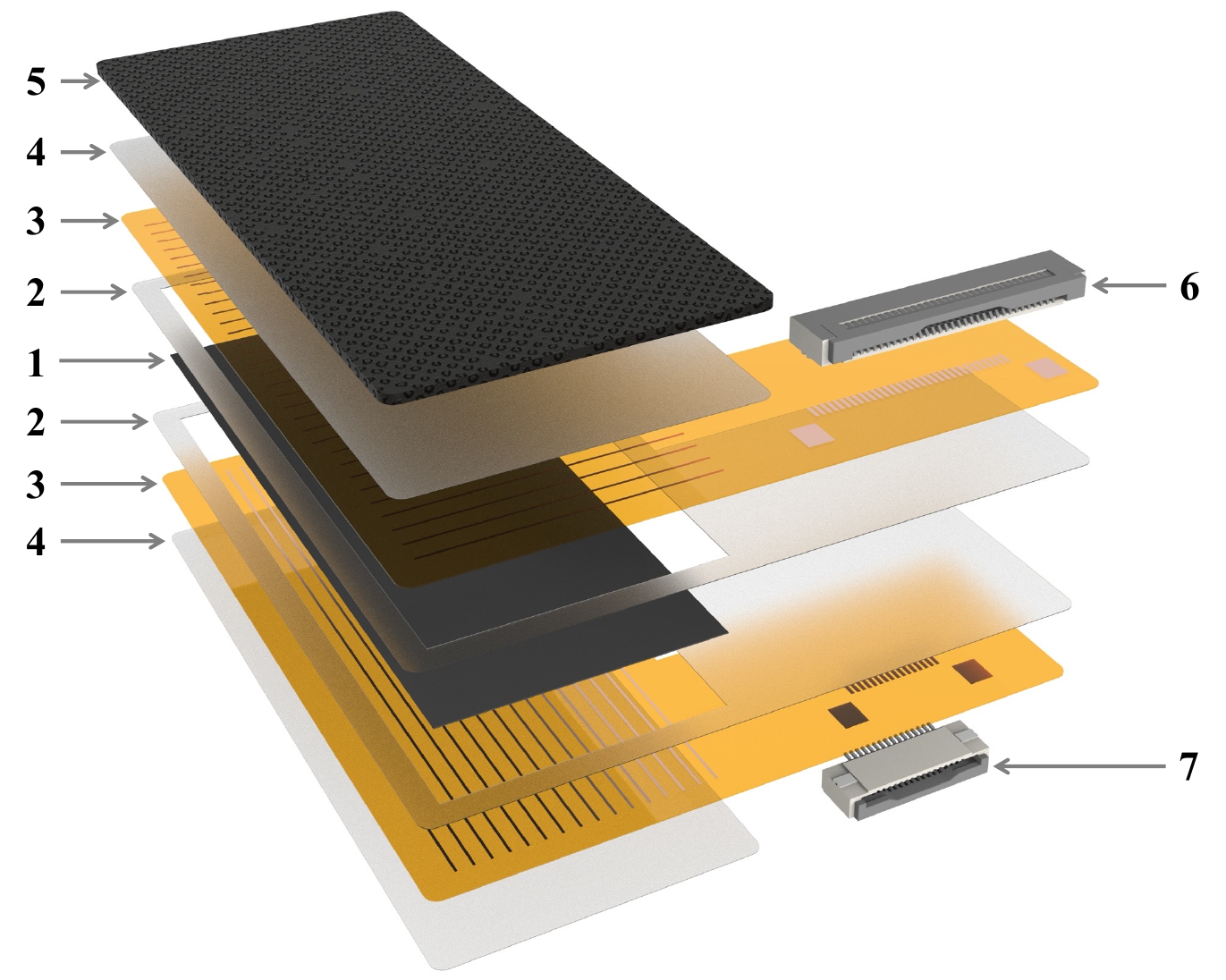}
    \vspace{-0.5cm}
    \caption{Tactile sensor structure of \sensor. Two FPCB-based conductive layers are stacked with a piezoresistive material to form a dense $16\times32$ tactile array over a $25\times60\,\mathrm{mm}$ sensing area. Numbers indicate key components: 1 -- piezoresistive layer, 2 -- 3M9077 tape, 3 -- conductive layer, 4 -- 3M468 tape, 5 -- cover layer, 6 -- 32-pin FFC connector, and 7 -- 16-pin FFC connector.}
    \label{fig:sensor}
\end{figure}

The tactile sensor structure is shown in Fig.~\ref{fig:sensor}. Each sensor is composed of two conductive layers with a piezoresistive material sandwiched between them, forming a dense $16\times32$ tactile array over a compact sensing area of $25\times60\,\mathrm{mm}$. This stacked structure provides a thin and lightweight sensing layer that can be directly integrated onto the robotic grippers.

For fabrication, we adopt the FPCB-based conductive-layer method of~\cite{murphy2025fits}. The two conductive layers are implemented using FPCBs rather than manually placed conductive cables. This design eliminates manual cable alignment and improves the consistency and simplicity of sensor assembly. In addition, the adhesive tapes and connectors are integrated into the FPCB design, eliminating manual tape trimming and soldering. As a result, the sensor side of the system becomes substantially easier to reproduce, complementing the improvements in electronics reproducibility described above.

\section{Experiments}

\begin{figure*}[th]
    \centering
    \includegraphics[width=\linewidth]{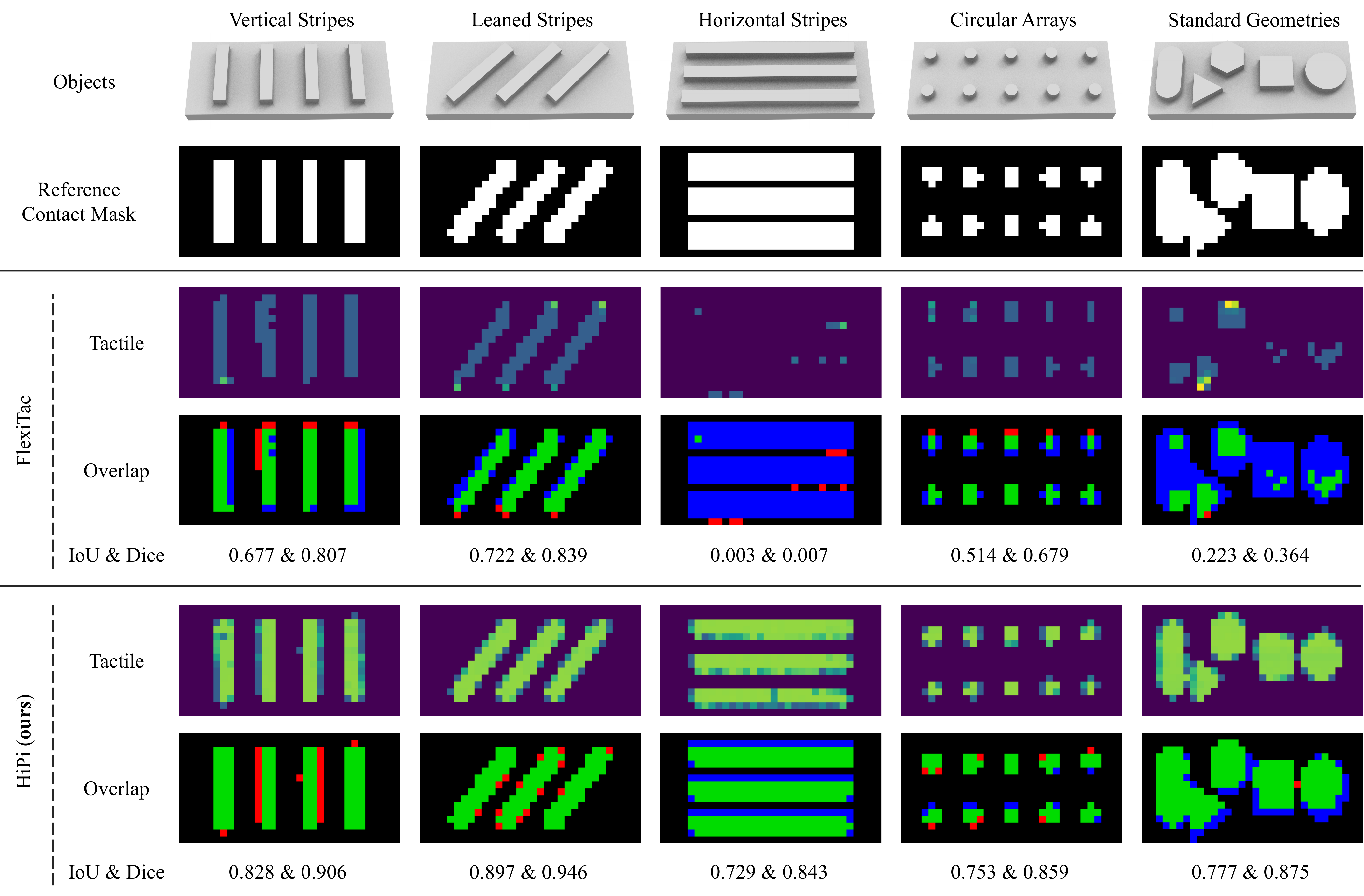}
    \vspace{-0.5cm}
    \caption{Comparison of tactile sensing fidelity between FlexiTac~\cite{huang2026flexitac} and \sensor. Five categories of 3D-printed contact patterns are used: vertical stripes, leaned stripes, horizontal stripes, circular arrays, and standard geometries. For each pattern, we show the object, the reference contact mask, the measured tactile response, the overlap visualization, and the corresponding IoU and Dice scores. In the overlap images, green indicates correctly recovered contact, blue indicates missed contact, and red indicates redundant contact.}
    \label{fig:exp_tactile_sensing}
\end{figure*}

We evaluate the high-fidelity sensing capability of \sensor with a focus on its low-crosstalk behavior. The goal is to test whether the system can preserve clear and localized contact patterns under structured contacts, rather than producing blurred or ghost responses caused by inter-taxel interference.

As shown in Fig.~\ref{fig:exp_tactile_sensing}, we design five categories of 3D-printed contact patterns: vertical stripes, leaned stripes, horizontal stripes, circular arrays, and standard geometries including a circle, triangle, square, hexagon, and straight slot. These patterns cover both repeated local structures and distinct geometric boundaries, providing a simple but informative benchmark for tactile image fidelity. For each test pattern, we generate a binary reference mask and compare it with the measured tactile response both qualitatively and quantitatively. Following~\cite{lin2025lighttact}, we report IoU and Dice to measure how well the recovered tactile pattern matches the target geometry.

Our main baseline is FlexiTac~\cite{huang2026flexitac}, which is open-source and highly reproducible. To isolate the effect of the readout electronics, the comparison in Fig.~\ref{fig:exp_tactile_sensing} uses the same tactile sensor structure in both cases and changes only the readout module. Across all five categories, \sensor consistently achieves higher IoU and Dice, indicating substantially better preservation of the intended contact geometry. Averaged over all patterns, FlexiTac achieves an IoU of $0.428$ and a Dice score of $0.539$, whereas \sensor achieves an IoU of $0.797$ and a Dice score of $0.886$.

A particularly revealing case is the horizontal-stripe pattern. \sensor recovers the three long stripe contacts with clear spatial separation, whereas FlexiTac almost entirely fails to preserve the contact layout, resulting in near-zero overlap scores. Similar degradation can also be observed in the standard-geometry case, where large connected contact regions lead to substantial loss of structure. More broadly, we observe that FlexiTac performs poorly when the contact area becomes large or when a large continuous region is activated along the signal-receiving direction. This behavior is consistent with crosstalk in its simplified readout circuits, where current leakage reduces signal locality and distorts the recovered pattern.

In contrast, \sensor maintains clear spatial structure across all examples, supporting that the proposed low-crosstalk readout design preserves contact geometry more faithfully. The remaining errors are comparatively minor and mainly appear near contact boundaries. These mismatches are more consistent with alignment uncertainty or fabrication variation than with severe signal corruption, since the overall tactile patterns remain sharp and geometrically consistent with the target masks. In addition, \sensor uses a single USB/MCU module to support $4\times16\times32$ taxels at $220\,\mathrm{Hz}$, whereas FlexiTac uses one USB/MCU module for $16\times32$ taxels at $100\,\mathrm{Hz}$. Overall, these results show that \sensor provides substantially higher-fidelity tactile readout than the baseline while also offering a more scalable architecture.


\section{Conclusion}

We presented \sensor, a reproducible high-fidelity piezoresistive sensing system for robotic manipulation. Building on low-crosstalk readout hardware, we redesigned the full system with practical reproducibility and deployability as primary objectives. The resulting system combines compact vendor-assembly-friendly electronics, a shared MCU architecture for multiple tactile modules, and FPCB-based sensor fabrication that eliminates manual cable alignment, tape trimming, and soldering. Quantitative experiments against FlexiTac show that \sensor preserves contact geometry much more faithfully across structured contact patterns, achieving substantially higher IoU and Dice while maintaining clear spatial separation under large or continuous contact regions. Together, these results show that \sensor bridges an important gap between reproducibility and signal quality in piezoresistive tactile sensing, providing an open and practical tactile hardware solution for robotic manipulation research.

\newpage
{
\footnotesize
\bibliographystyle{ieeetr}
\bibliography{main}
}

\end{document}